\documentclass[a4paper]{article}

\pdfoutput=1

\usepackage[english]{babel}
\usepackage[utf8x]{inputenc}
\usepackage[T1]{fontenc}
\usepackage{authblk}
\usepackage{subcaption}
\usepackage{verbatim}
\usepackage[font=small,labelfont=bf]{caption}

\usepackage[a4paper,top=3cm,bottom=2cm,left=3cm,right=3cm,marginparwidth=1.75cm]{geometry}

\usepackage{amsmath}
\usepackage{graphicx}
\usepackage[colorinlistoftodos]{todonotes}
\usepackage[colorlinks=true, allcolors=blue]{hyperref}
\usepackage{longtable}
\usepackage{float}

\newcommand{\beginsupplement}{%
        \setcounter{table}{0}
        \renewcommand{\thetable}{S\arabic{table}}%
        \setcounter{figure}{0}
        \renewcommand{\thefigure}{S\arabic{figure}}%
     }
     
\newcommand{\hlinetable}{%
	\begin{center}
	\vspace{10mm}
	\noindent\rule{3cm}{0.4pt}
	\vspace{10mm}
	\end{center}
    }

\title{When Will AI Exceed Human Performance? \\Evidence from AI Experts}

\author[1,2]{Katja Grace}
\author[2]{John Salvatier}
\author[1,3]{Allan Dafoe}
\author[3]{Baobao Zhang}
\author[1]{Owain Evans}
\affil[1]{Future of Humanity Institute, Oxford University} 
\affil[2]{AI Impacts}
\affil[3]{Department of Political Science, Yale University}

\date{\vspace{-5ex}}
\begin{document}
\maketitle

\begin{abstract}
\noindent Advances in artificial intelligence (AI) will transform modern life by reshaping transportation, health, science, finance, and the military \cite{stone2016artificial,domingos2015master,bostrom2014superintelligence}. To adapt public policy, we need to better anticipate these advances \cite{brynjolfsson2014second,calo2015robotics}. Here we report the results from a large survey of machine learning researchers on their beliefs about progress in AI. Researchers predict AI will outperform humans in many activities in the next ten years, such as translating languages (by 2024), writing high-school essays (by 2026), driving a truck (by 2027), working in retail (by 2031), writing a bestselling book (by 2049), and working as a surgeon (by 2053). Researchers believe there is a 50\% chance of AI outperforming humans in all tasks in 45 years and of automating all human jobs in 120 years, with Asian respondents expecting these dates much sooner than North Americans. These results will inform discussion amongst researchers and policymakers about anticipating and managing trends in AI.
\end{abstract}

\noindent 
\subsection*{Introduction}
Advances in artificial intelligence (AI) will have massive social consequences. Self-driving technology might replace millions of driving jobs over the coming decade. In addition to possible unemployment, the transition will bring new challenges, such as rebuilding infrastructure, protecting vehicle cyber-security, and adapting laws and regulations \cite{calo2015robotics}. New challenges, both for AI developers and policy-makers, will also arise from applications in law enforcement, military technology, and marketing \cite{jiang2015self}. To prepare for these challenges, accurate forecasting of transformative AI would be invaluable. 

Several sources provide objective evidence about future AI advances: trends in computing hardware \cite{nordhaus2007two}, task performance \cite{grace2013algorithmic}, and the automation of labor \cite{brynjolfsson2012race}. The predictions of AI experts provide crucial additional information \cite{baum2011long,muller2016future, walsh2017expert}. We survey a large, representative sample of AI experts. Our questions cover the timing of AI advances (including both practical applications of AI and the automation of various human jobs), as well as the social and ethical impacts of AI.

\subsection*{Survey Method}

Our survey population was all researchers who published at the 2015 NIPS and ICML conferences (two of the premier venues for peer-reviewed research in machine learning). A total of 352 researchers responded to our survey invitation (21\% of the 1634 authors we contacted). Our questions concerned the timing of specific AI capabilities (e.g. folding laundry, language translation), superiority at specific occupations (e.g. truck driver, surgeon), superiority over humans at all tasks, and the social impacts of advanced AI. See \hyperref[content]{Survey Content} for details.

\subsection*{Time Until Machines Outperform Humans}\label{automation}

AI would have profound social consequences if all tasks were more cost effectively accomplished by machines. Our survey used the following definition:

\begin{quote}
``High-level machine intelligence'' (HLMI) is achieved when unaided machines can accomplish every task better and more cheaply than human workers.
\end{quote}

\begin{figure}[H]
\includegraphics[width=\textwidth]{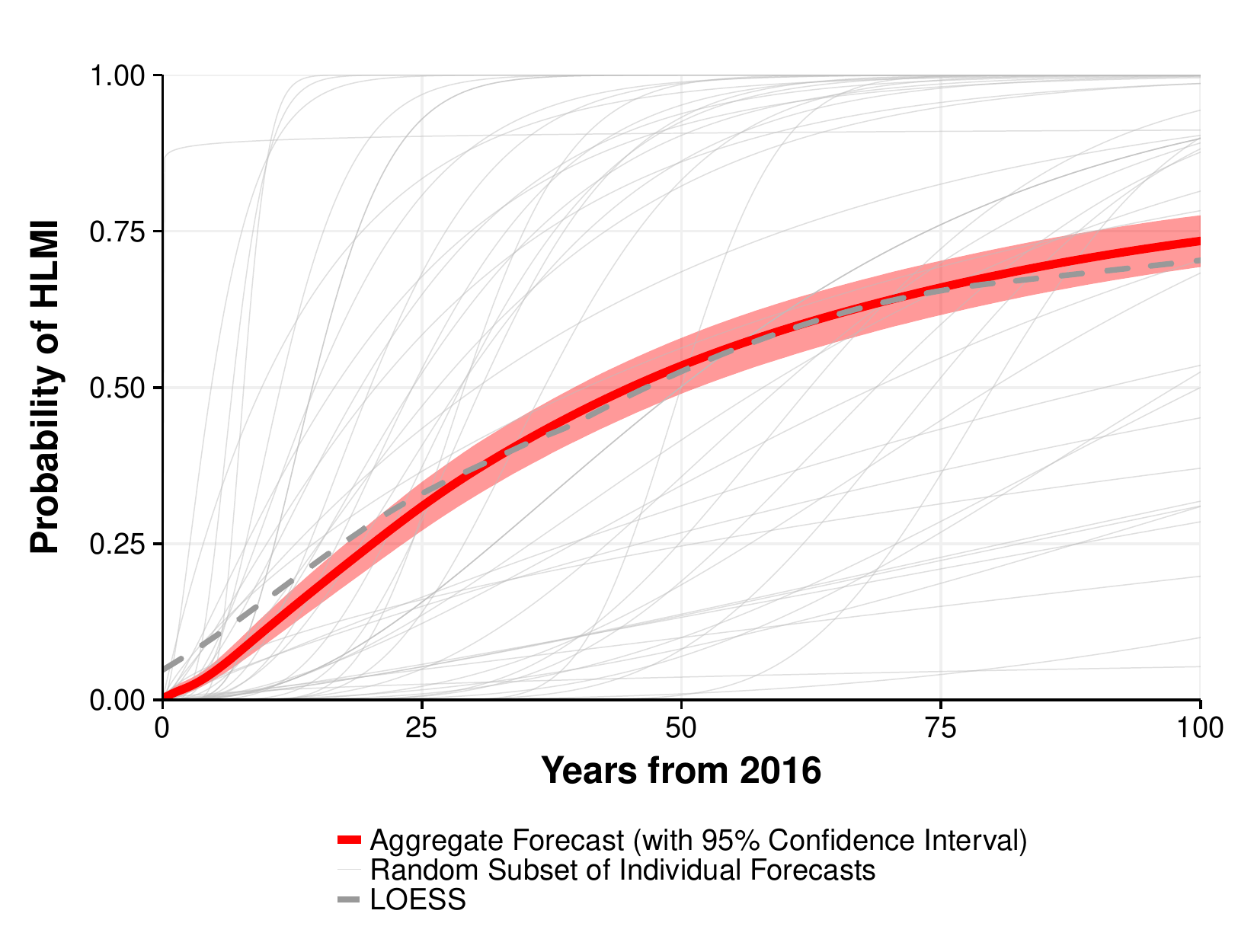}
\caption{\textbf{Aggregate subjective probability of `high-level machine intelligence’ arrival by future years}. Each respondent provided three data points for their forecast and these were fit to the Gamma CDF by least squares to produce the grey CDFs. The ``Aggregate Forecast'' is the mean distribution over all individual CDFs (also called the “mixture” distribution). The confidence interval was generated by bootstrapping (clustering on respondents) and plotting the 95\% interval for estimated probabilities at each year. The LOESS curve is a non-parametric regression on all data points.
}
\label{fig:cdf}
\end{figure}

\noindent Each individual respondent estimated the probability of HLMI arriving in future years. Taking the mean over each individual, the aggregate forecast gave a 50\% chance of HLMI occurring within 45 years and a 10\% chance of it occurring within 9 years. Figure~\ref{fig:cdf} displays the probabilistic predictions for a random subset of individuals, as well as the mean predictions. There is large inter-subject variation: Figure~\ref{fig:region} shows that Asian respondents expect HLMI in 30 years, whereas North Americans expect it in 74 years.

\noindent While most participants were asked about HLMI, a subset were asked a logically similar question that emphasized consequences for employment. The question defined full automation of labor as:

\begin{quote}
when all occupations are fully automatable. That is, when for any occupation, machines could be built to carry out the task better and more cheaply than human workers.
\end{quote}

\noindent Forecasts for full automation of labor were much later than for HLMI: the mean of the individual beliefs assigned a 50\% probability in 122 years from now and a 10\% probability in 20 years. 

\begin{figure}[H] 
\includegraphics[width=\textwidth]{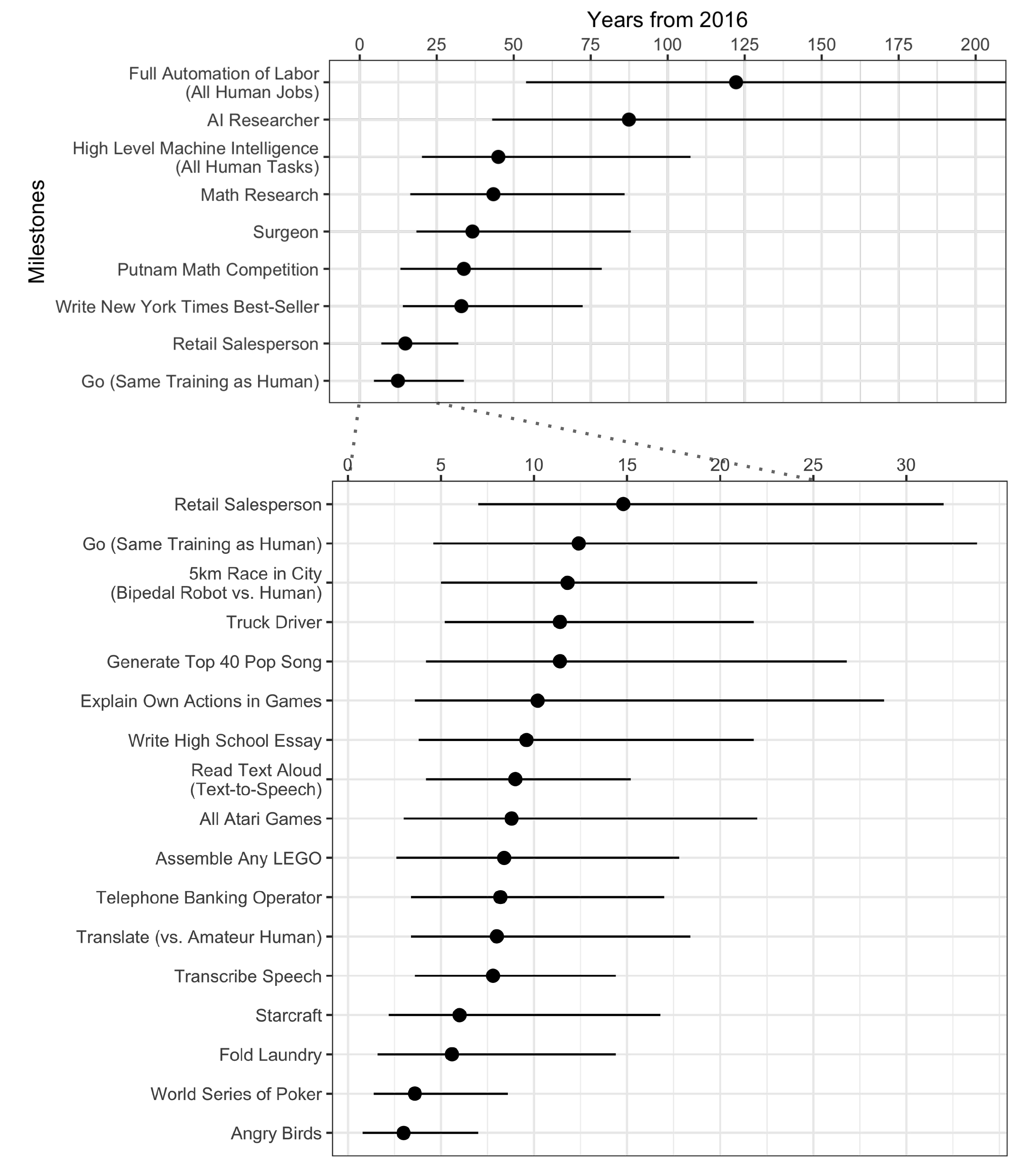}
\caption{\textbf{Timeline of Median Estimates (with 50\% intervals) for AI Achieving Human Performance}. Timelines showing 50\% probability intervals for achieving selected AI milestones. Specifically, intervals represent the date range from the 25\% to 75\% probability of the event occurring, calculated from the mean of individual CDFs as in Fig. \ref{fig:cdf}. Circles denote the 50\%-probability year. Each milestone is for AI to achieve or surpass human expert/professional performance (full descriptions in Table \ref{tab:s5}). Note that these intervals represent the uncertainty of survey respondents, not estimation uncertainty.}
\label{fig:timeline}
\end{figure}

\noindent Respondents were also asked when 32 “milestones” for AI would become feasible. The full descriptions of the milestone are in Table \ref{tab:s5}. Each milestone was considered by a random subset of respondents (n≥24). Respondents expected (mean probability of 50\%) 20 of the 32 AI milestones to be reached within ten years. Fig. \ref{fig:timeline} displays timelines for a subset of milestones.

\subsection*{Intelligence Explosion, Outcomes, AI Safety}

The prospect of advances in AI raises important questions. Will progress in AI become explosively fast once AI research and development itself can be automated? How will high-level machine intelligence (HLMI) affect economic growth? What are the chances this will lead to extreme outcomes (either positive or negative)? What should be done to help ensure AI progress is beneficial? Table \ref{tab:s4} displays results for questions we asked on these topics. Here are some key findings:

\begin{enumerate}
\item \textbf{Researchers believe the field of machine learning has accelerated in recent years.} We asked researchers whether the rate of progress in machine learning was faster in the first or second half of their career. Sixty-seven percent (67\%) said progress was faster in the second half of their career and only 10\% said progress was faster in the first half. The median career length among respondents was 6 years.

\item \textbf{Explosive progress in AI after HLMI is seen as possible but improbable.} Some authors have argued that once HLMI is achieved, AI systems will quickly become vastly superior to humans in all tasks \cite{bostrom2014superintelligence,good1966speculations}. This acceleration has been called the ``intelligence explosion.'' We asked respondents for the probability that AI would perform vastly better than humans in all tasks two years after HLMI is achieved. The median probability was 10\% (interquartile range: 1-25\%). We also asked respondents for the probability of explosive global technological improvement two years after HLMI. Here the median probability was 20\% (interquartile range 5-50\%).

\item \textbf{HLMI is seen as likely to have positive outcomes but catastrophic risks are possible.} Respondents were asked whether HLMI would have a positive or negative impact on humanity over the long run. They assigned probabilities to outcomes on a five-point scale. The median probability was 25\% for a ``good'' outcome and 20\% for an ``extremely good'' outcome. By contrast, the probability was 10\% for a bad outcome and 5\% for an outcome described as ``Extremely Bad (e.g., human extinction).''

\item \textbf{Society should prioritize research aimed at minimizing the potential risks of AI.} Forty-eight percent of respondents think that research on minimizing the risks of AI should be prioritized by society more than the status quo (with only 12\% wishing for less).

\end{enumerate}

\begin{figure}[H]
\centering
\includegraphics[width=0.8\linewidth]{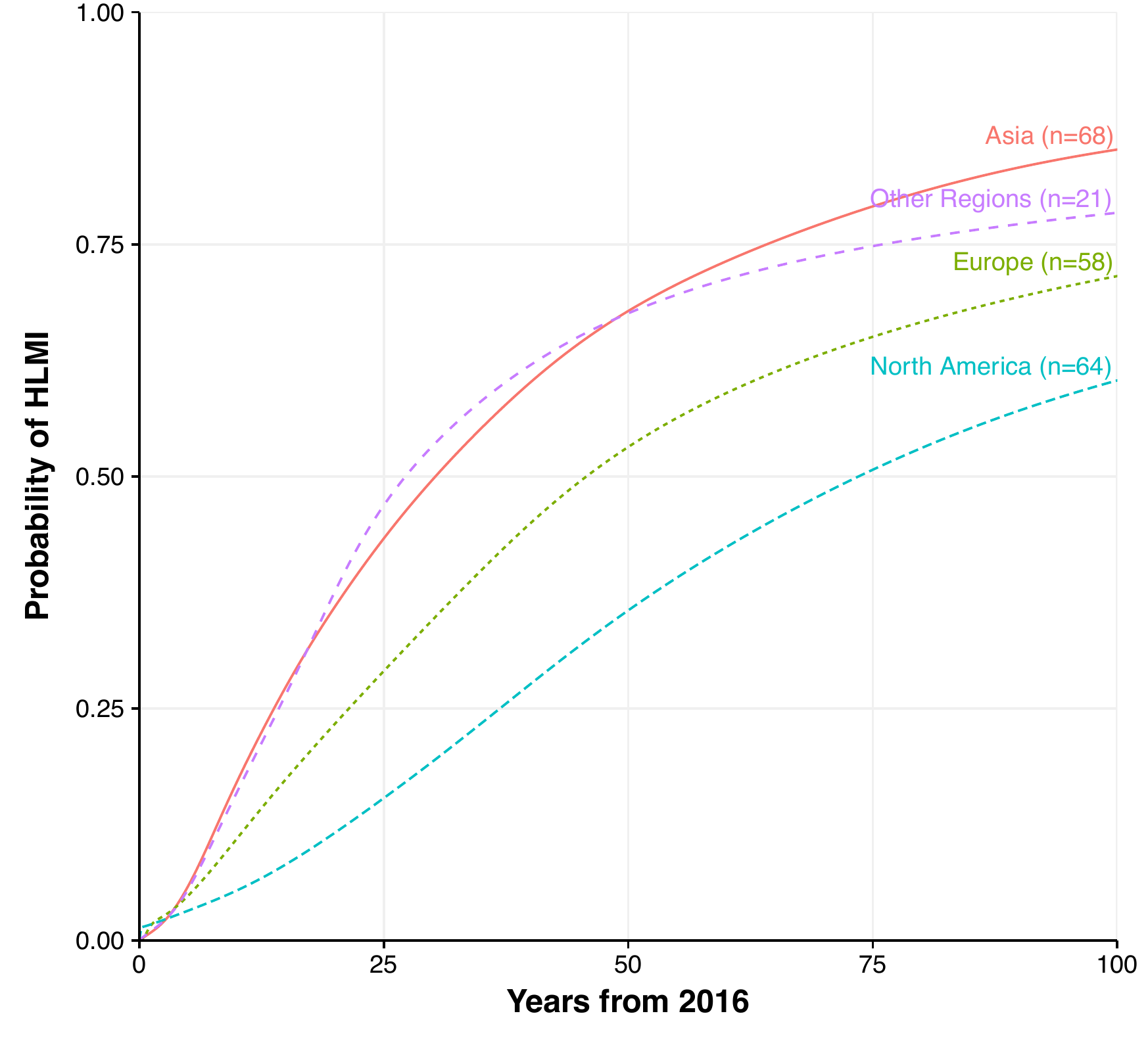}
\caption{\textbf{Aggregate Forecast (computed as in Figure 1) for HLMI, grouped by region in which respondent was an undergraduate.} Additional regions (Middle East, S. America, Africa, Oceania) had much smaller numbers and are grouped as ``Other Regions.''} 
\label{fig:region}
\end{figure}

\subsection*{Asians expect HLMI 44 years before North Americans}

\begin{figure}[H]
\begin{subfigure}{.5\textwidth}
  \centering
    \caption{Top 4 Undergraduate Country HLMI CDFs}
  \includegraphics[width=.99\linewidth]{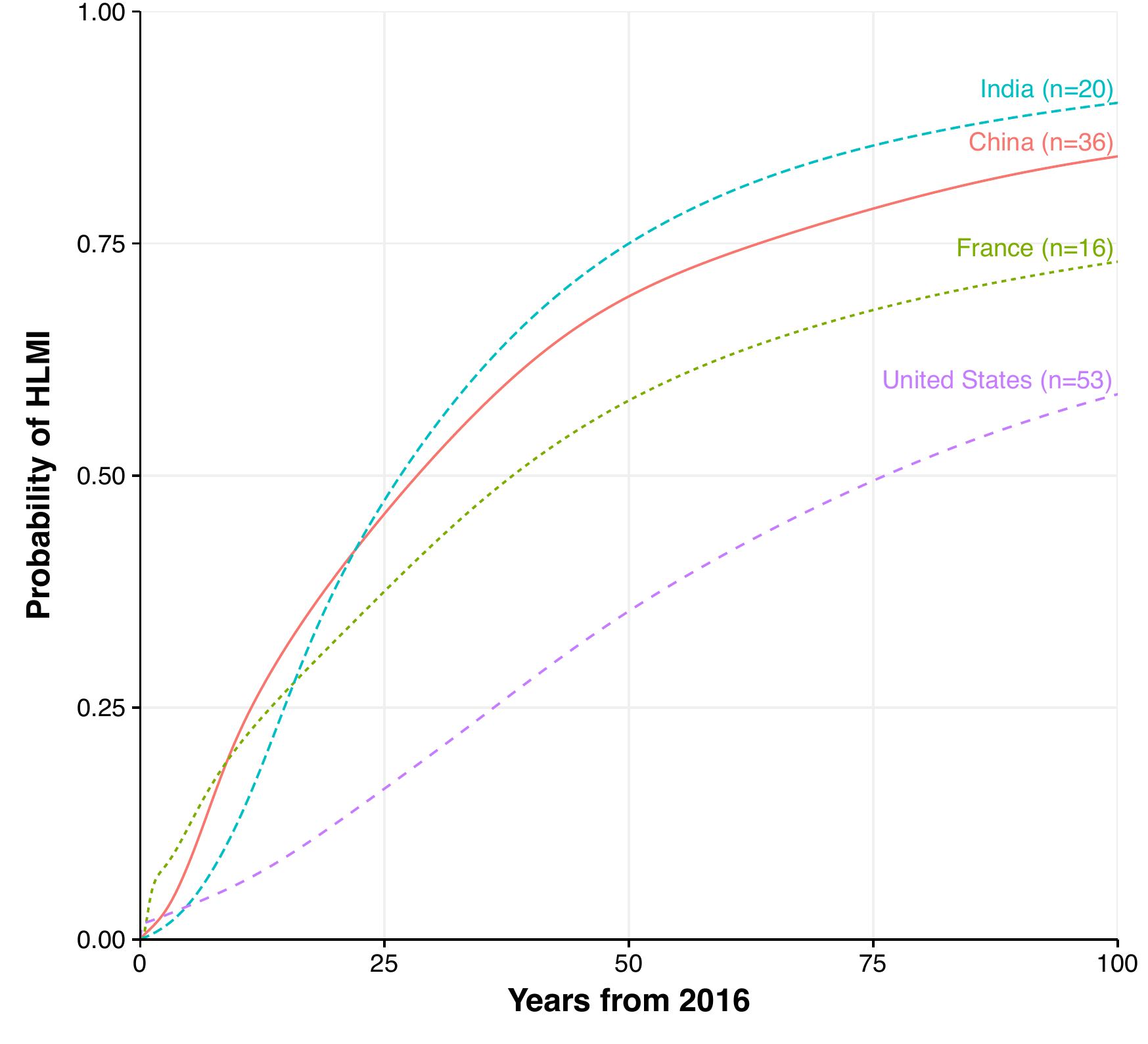}
  \label{fig:s1fig1}
\end{subfigure}%
\begin{subfigure}{.5\textwidth}
  \centering
    \caption{Time in Field Quantile HLMI CDFs}
  \includegraphics[width=.99\linewidth]{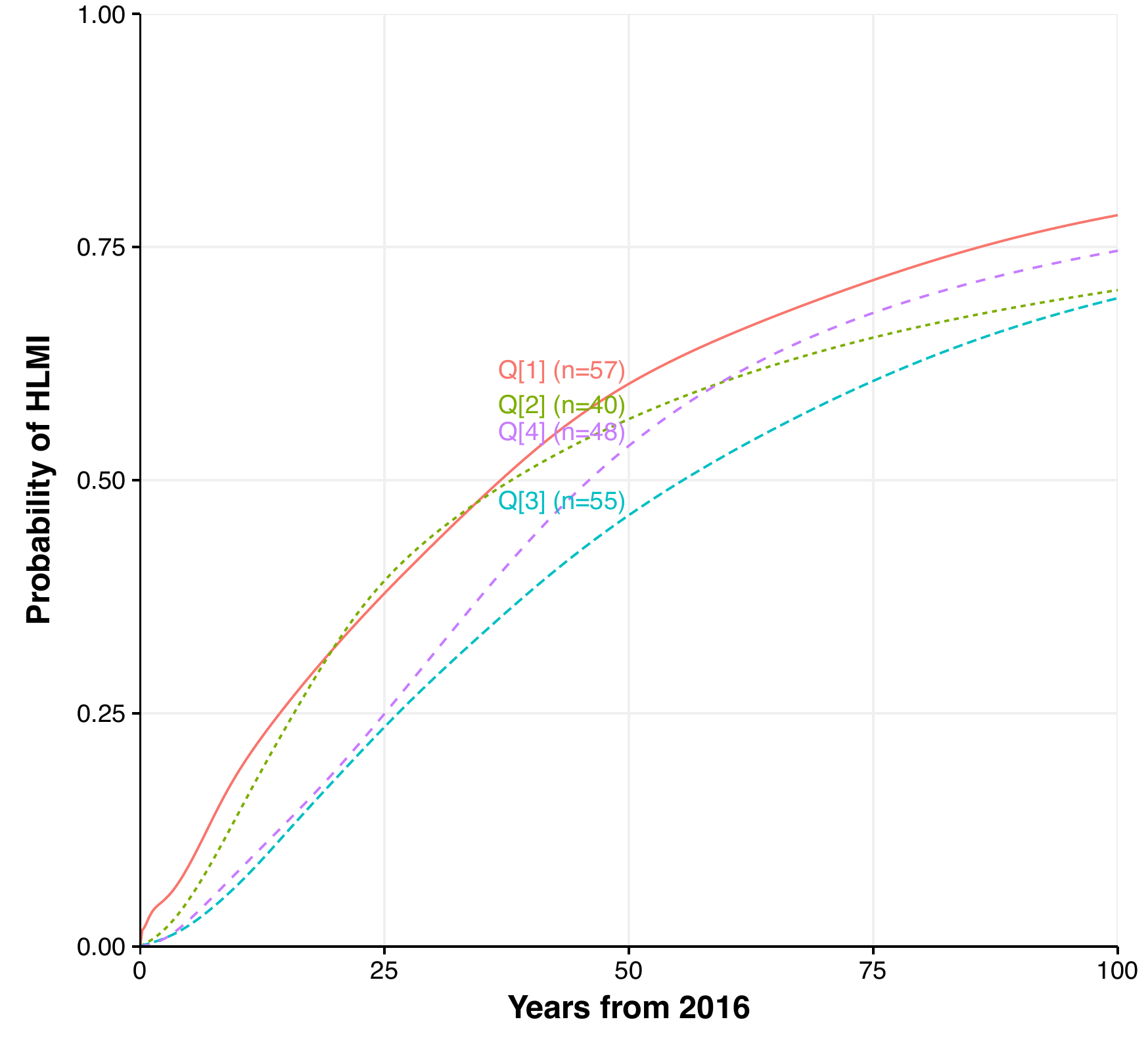}
  \label{fig:s1fig2}
\end{subfigure}
\begin{subfigure}{.5\textwidth}
  \centering
    \caption{Citation Count Quartile HLMI CDFs}
  \includegraphics[width=.99\linewidth]{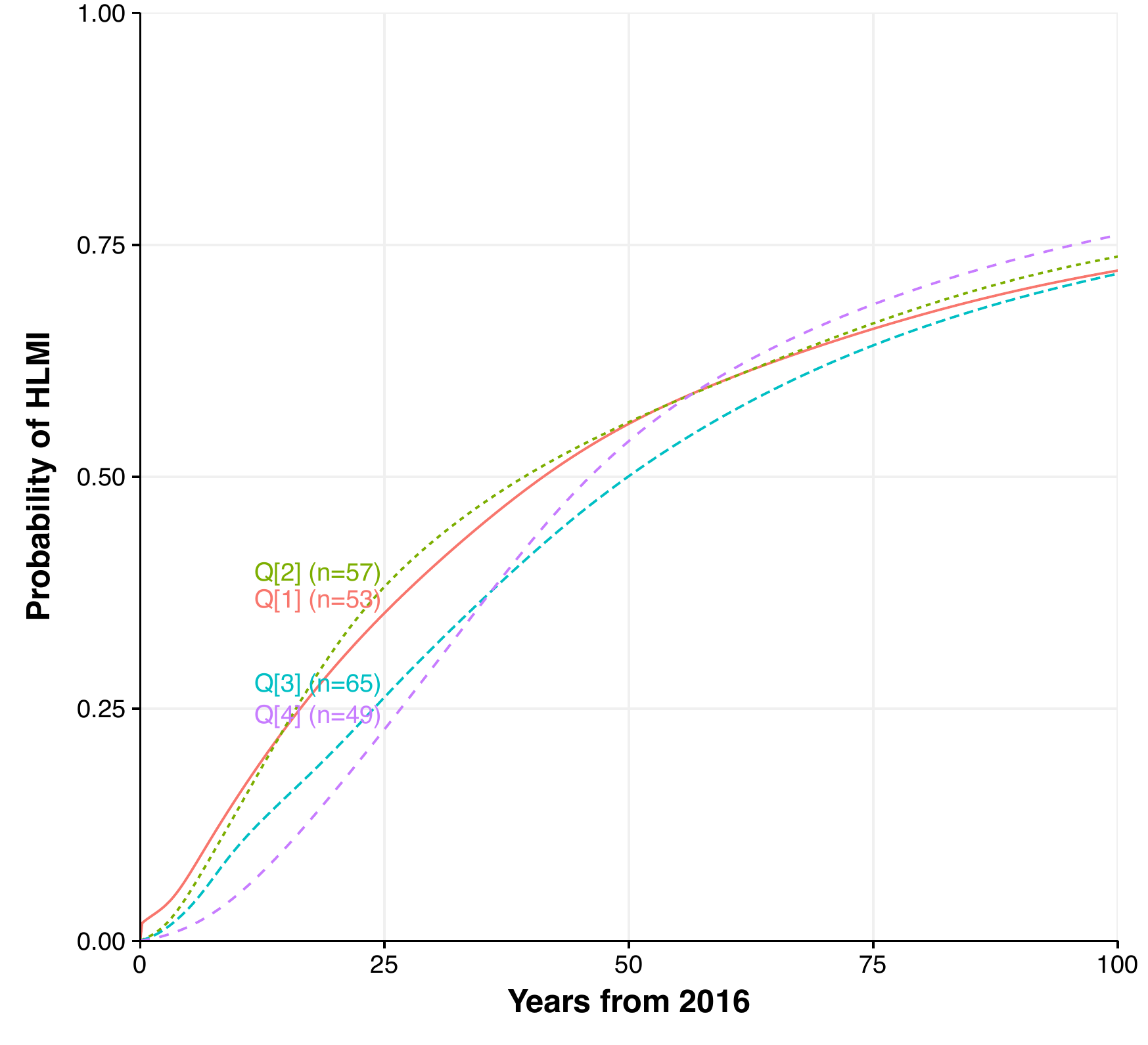}
  \label{fig:s1fig3}
\end{subfigure}
\caption{\textbf{Aggregate subjective probability of HLMI arrival by demographic group.} Each graph curve is an Aggregate Forecasts CDF, computed using the procedure described in Figure 1 and in ``Elicitation of Beliefs.'' Figure \ref{fig:s1fig1}  shows aggregate HLMI predictions for the four countries with the most respondents in our survey. Figure \ref{fig:s1fig2} shows predictions grouped by quartiles for seniority (measured by time since they started a PhD). Figure \ref{fig:s1fig3} shows predictions grouped by quartiles for citation count. ``Q4'' indicates the top quartile (i.e. the most senior researchers or the researchers with most citations). }
\label{fig:s_cdf}
\end{figure}

Figure 3 shows big differences between individual respondents in when they predict HLMI will arrive. Both citation count and seniority were not predictive of HLMI timelines (see Fig. \ref{fig:s_cdf} and the results of a regression in Table \ref{tab:s2}). However, respondents from different regions had striking differences in HLMI predictions. Fig. \ref{fig:region} shows an aggregate prediction for HLMI of 30 years for Asian respondents and 74 years for North Americans. Fig. \ref{fig:s_cdf} displays a similar gap between the two countries with the most respondents in the survey: China (median 28 years) and USA (median 76 years). Similarly, the aggregate year for a 50\% probability for automation of each job we asked about (including truck driver and surgeon) was predicted to be earlier by Asians than by North Americans (Table \ref{tab:s2}). Note that we used respondents’ undergraduate institution as a proxy for country of origin and that many Asian respondents now study or work outside Asia.

\subsection*{Was our sample representative?}

One concern with any kind of survey is non-response bias; in particular, researchers with strong views may be more likely to fill out a survey. We tried to mitigate this effect by making the survey short (12 minutes) and confidential, and by not mentioning the survey’s content or goals in our invitation email. Our response rate was 21\%. To investigate possible non-response bias, we collected demographic data for both our respondents (n=406) and a random sample (n=399) of NIPS/ICML researchers who did not respond. Results are shown in Table \ref{tab:s3}. Differences between the groups in citation count, seniority, gender, and country of origin are small. While we cannot rule out non-response biases due to unmeasured variables, we can rule out large bias due to the demographic variables we measured. Our demographic data also shows that our respondents included many highly-cited researchers (mostly in machine learning but also in statistics, computer science theory, and neuroscience) and came from 43 countries (vs. a total of 52 for everyone we sampled). A majority work in academia (82\%), while 21\% work in industry.

A second concern is that NIPS and ICML authors are representative of \emph{machine learning} but not of the field of artificial intelligence as a whole. This concern could be addressed in future work by surveying a broader range of experts across computer science, robotics, and the cognitive sciences. In fact, a 2017 survey by Walsh \cite{walsh2017expert} asked a broad range of AI and robotics experts a question about HLMI almost identical to ours. For a 50\% chance of HLMI, the median prediction in this survey was 2065 for robiticists and 2061 for AI experts. Our machine learning experts predicted 2057. This is very close Walsh's results and suggests that our conclusions about expert views on HLMI are robust to surveying experts outside machine learning.\footnote{The difference in medians between us and Walsh is tiny compared to differences between Asians and North Americans in our study and does not provide evidence of a substantial difference between groups of experts.} It's still possible that groups of experts differ on topics other than HLMI timelines.

\subsection*{Discussion}

Why think AI experts have any ability to foresee AI progress? In the domain of political science, a long-term study found that experts were worse than crude statistical extrapolations at predicting political outcomes \cite{tetlock2005expert}. AI progress, which relies on scientific breakthroughs, may appear intrinsically harder to predict. Yet there are reasons for optimism. While individual breakthroughs are unpredictable, longer term progress in R\&D for many domains (including computer hardware, genomics, solar energy) has been impressively regular \cite{farmer2016predictable}. Such regularity is also displayed by trends \cite{grace2013algorithmic} in AI performance in SAT problem solving, games-playing, and computer vision and could be exploited by AI experts in their predictions. Finally, it is well established that aggregating individual predictions can lead to big improvements over the predictions of a random individual \cite{ungar2012good}. Further work could use our data to make optimized forecasts. Moreover, many of the AI milestones (Fig. \ref{fig:timeline}) were forecast to be achieved in the next decade, providing ground-truth evidence about the reliability of individual experts.

\newpage
\bibliographystyle{unsrt}
\bibliography{my_bib}

\newpage
\section*{Supplementary Information}
\beginsupplement

\subsection*{Survey Content}\label{content}

We developed questions through a series of interviews with Machine Learning researchers. Our survey questions were as follows:
\begin{enumerate}
\item Three sets of questions eliciting HLMI predictions by different framings: asking directly about HLMI, asking about the automatability of all human occupations, and asking about recent progress in AI from which we might extrapolate. 
\item Three questions about the probability of an ``intelligence explosion''.
\item One question about the welfare implications of HLMI.
\item A set of questions about the effect of different inputs on the rate of AI research (e.g., hardware progress).
\item Two questions about sources of disagreement about AI timelines and ``AI Safety.''
\item Thirty-two questions about when AI will achieve narrow “milestones”. 
\item Two sets of questions on AI Safety research: one about AI systems with non-aligned goals, and one on the prioritization of Safety research in general.
\item A set of demographic questions, including ones about how much thought respondents have given to these topics in the past.
The questions were asked via an online Qualtrics survey. (The Qualtrics file will be shared to enable replication.) Participants were invited by email and were offered a financial reward for completing the survey. Questions were asked in roughly the order above and respondents received a randomized subset of questions. Surveys were completed between May 3rd 2016 and June 28th 2016.
\end{enumerate}

\noindent Our goal in defining “high-level machine intelligence” (HLMI) was to capture the widely-discussed notions of “human-level AI” or “general AI” (which contrasts with “narrow AI”) \cite{bostrom2014superintelligence}. We consulted all previous surveys of AI experts and based our definition on that of an earlier survey \cite{muller2016future}. Their definition of HLMI was a machine that “can carry out most human professions at least as well as a typical human.” Our definition is more demanding and requires machines to be better at all tasks than humans (while also being more cost-effective). Since earlier surveys often use less demanding notions of HLMI, they should (all other things being equal) predict earlier arrival for HLMI.

\subsection*{Demographic Information}

The demographic information on respondents and non-respondents (Table \ref{tab:s3}) was collected from public sources, such as academic websites, LinkedIn profiles, and Google Scholar profiles. Citation count and seniority (i.e. numbers of years since the start of PhD) were collected in February 2017.

\subsection*{Elicitation of Beliefs}\label{elicitation}

\begin{figure}[H]
\includegraphics[width=\textwidth]{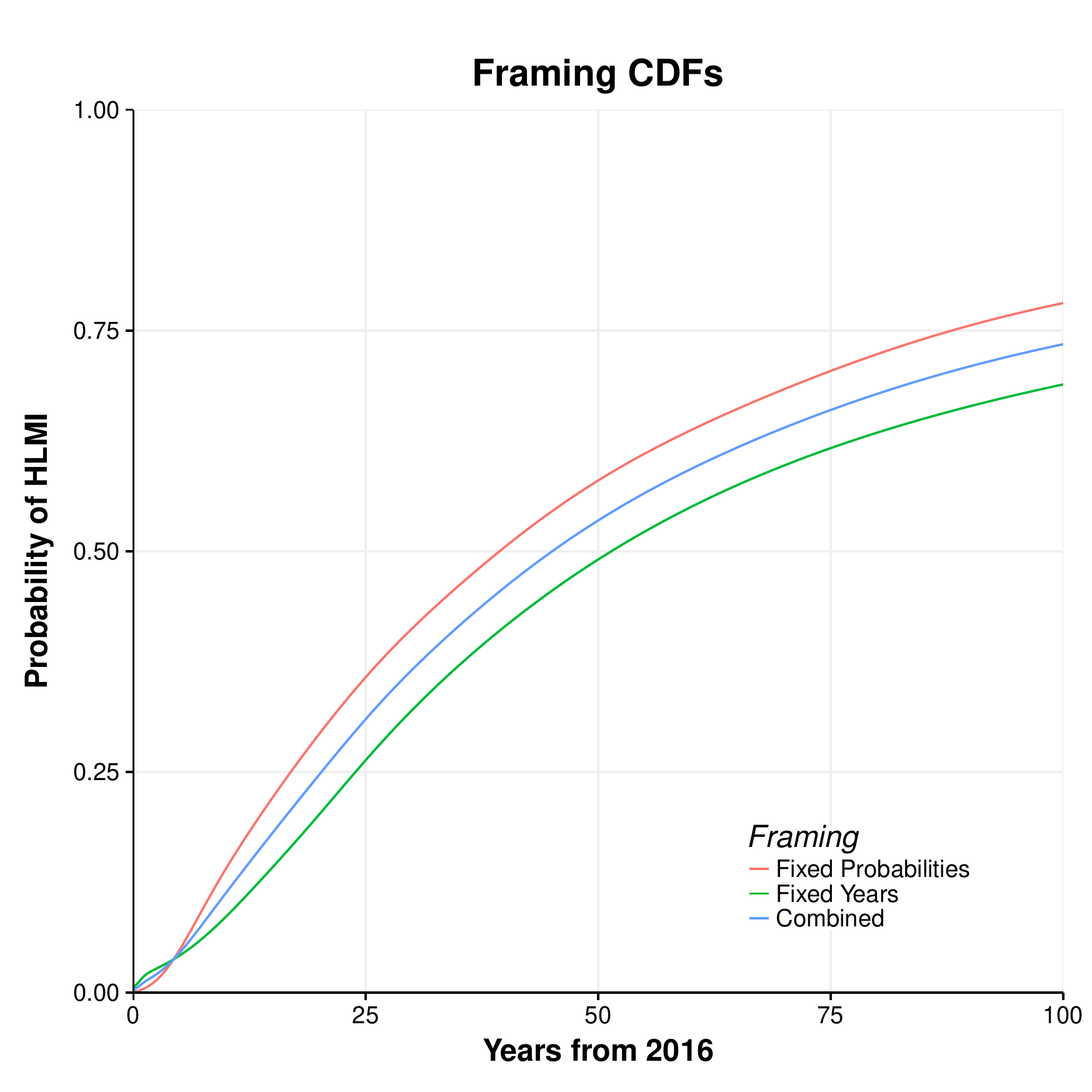}
\caption{\textbf{Aggregate subjective probability of HLMI arrival for two framings of the question.} The ``fixed probabilities'' and ``fixed years'' curves are each an aggregate forecast for HLMI predictions, computed using the same procedure as in Fig.~\ref{fig:cdf}. These two framings of questions about HLMI are explained in ``Elicitation of Beliefs'' above. The ``combined'' curve is an average over these two framings and is the curve used in Fig.~\ref{fig:cdf}.}
\label{fig:s_framing}
\end{figure}

Many of our questions ask when an event will happen. For prediction tasks, ideal Bayesian agents provide a cumulative distribution function (CDF) from time to the cumulative probability of the event. When eliciting points on respondents’ CDFs, we framed questions in two different ways, which we call “fixed-probability” and “fixed-years”. Fixed-probability questions ask by which year an event has an p\% cumulative probability (for p=10\%, 50\%, 90\%). Fixed-year questions ask for the cumulative probability of the event by year y (for y=10, 25, 50). The former framing was used in recent surveys of HLMI timelines; the latter framing is used in the psychological literature on forecasting \cite{tidwell2013eliciting,wallsten2016efficiently}. With a limited question budget, the two framings will sample different points on the CDF; otherwise, they are logically equivalent. Yet our survey respondents do not treat them as logically equivalent. We observed effects of question framing in all our prediction questions, as well as in pilot studies. Differences in these two framings have previously been documented in the forecasting literature \cite{tidwell2013eliciting,wallsten2016efficiently} but there is no clear guidance on which framing leads to more accurate predictions. Thus we simply average over the two framings when computing CDF estimates for HLMI and for tasks. HLMI predictions for each framing are shown in Fig. \ref{fig:s_framing}.

\subsection*{Statistics}

For each timeline probability question (see Figures 1 and 2), we computed an aggregate distribution by fitting a gamma CDF to each individual’s responses using least squares and then taking the mixture distribution of all individuals. Reported medians and quantiles were computed on this summary distribution. The confidence intervals were generated by bootstrapping (clustering on respondents with 10,000 draws) and plotting the 95\% interval for estimated probabilities at each year.
The time-in-field and citations comparisons between respondents and non-respondents (Table \ref{tab:s3}) were done using two-tailed t-tests. The region and gender proportions were done using two-sided proportion tests.
The significance test for the effect of region on HLMI date (Table \ref{tab:s2}) was done using robust linear regression using the R function \texttt{rlm} from the MASS package to do the regression and then the \texttt{f.robtest} function from the \texttt{sfsmisc} package to do a robust F-test significance. 

\newpage
\section*{Supplementary Figures}

\newpage
\section*{Supplementary Tables}

\subsection*{S1: Automation Predictions by Researcher Region}
This question asked when automation of the job would become feasible, and cumulative probabilities were elicited as in the HLMI and milestone prediction questions. The definition of “full automation” is given above (p.\pageref{automation}). For the “NA/Asia gap”, we subtract the Asian from the N. American median estimates.

\begin{table}[H]
\centering
\caption{Median estimate (in years from 2016) for automation of human jobs by region of undergraduate institution \vspace{3mm}}
\label{tab:s1}
\begin{tabular}{|l|l|l|l|l|}
\hline
\textbf{Question}  & \textbf{Europe} & \textbf{N. America} & \textbf{Asia} & \textbf{NA/Asia gap} \\ \hline
Full Automation    & 130.8           & 168.6               & 104.2         & +64.4                               \\ \hline
Truck Driver       & 13.2            & 10.6                & 10.2          & +0.4                                \\ \hline
Surgeon       	   & 46.4            & 41.0                & 31.4          & +9.6                                \\ \hline
Retail Salesperson & 18.8            & 20.2                & 10.0          & +10.2                               \\ \hline
AI Researcher      & 80.0            & 123.6               & 109.0         & +14.6                               \\ \hline
\end{tabular}
\end{table}

\hlinetable
\subsection*{S2: Regression of HLMI Prediction on Demographic Features}
We standardized inputs and regressed the log of the median years until HLMI for respondents on gender, log of citations, seniority (i.e. numbers of years since start of PhD), question framing (``fixed-probability'' vs. ``fixed-years'') and region where the individual was an undergraduate. We used a robust linear regression.

\begin{table}[H]
\caption{Robust linear regression for individual HLMI predictions \vspace{2mm}}
\centering
\label{tab:s2}
\begin{tabular}{|l|l|l|l|l|p{1.5cm}|}
\hline
\textbf{term}                    & \textbf{Estimate} & \textbf{SE} & \textbf{\textit{t}-statistic} & \textbf{\textit{p}-value} & \textbf{Wald \textit{F}-statistic} \\ \hline
(Intercept)                      & 3.65038           & 0.17320     & 21.07635             & 0.00000          & 458.0979                  \\ \hline
Gender = ``female''                  & -0.25473          & 0.39445     & -0.64578             & 0.55320          & 0.3529552                 \\ \hline
log(citation\_count)             & -0.10303          & 0.13286     & -0.77546             & 0.44722          & 0.5802456                 \\ \hline
Seniority (years)                & 0.09651           & 0.13090     & 0.73728              & 0.46689          & 0.5316029                 \\ \hline
Framing = ``fixed\_probabilities'' & -0.34076          & 0.16811     & -2.02704             & 0.04414          & 4.109484                  \\ \hline
Region = ``Europe''                & 0.51848           & 0.21523     & 2.40898              & 0.01582          & 5.93565                   \\ \hline
Region = ``M.East''                & -0.22763          & 0.37091     & -0.61369             & 0.54430          & 0.3690532                 \\ \hline
Region = ``N.America''             & 1.04974           & 0.20849     & 5.03496              & 0.00000          & 25.32004                  \\ \hline
Region = ``Other''                 & -0.26700          & 0.58311     & -0.45788             & 0.63278          & 0.2291022                 \\ \hline
\end{tabular}
\end{table}

\newpage
\subsection*{S3: Demographics of Respondents vs. Non-respondents}
There were (n=406) respondents and (n=399) non-respondents. Non-respondents were randomly sampled from all NIPS/ICML authors who did not respond to our survey invitation. Subjects with missing data for region of undergraduate institution or for gender are grouped in “NA”. Missing data for citations and seniority is ignored in computing averages. Statistical tests are explained in section “Statistics” above. 

\vspace{6mm}

\begin{table}[H]
\caption{Demographic differences between respondents and non-respondents}
\label{tab:s3}
\begin{tabular}{|p{3cm}|p{3cm}|p{3cm}|l|}
\hline
\textbf{Undergraduate region} & \textbf{Respondent proportion} & \textbf{Non-respondent proportion} & \textbf{p-test \textit{p}-value} \\ \hline
Asia                          & 0.305                          & 0.343                              & 0.283                   \\ \hline
Europe                        & 0.271                          & 0.236                              & 0.284                   \\ \hline
Middle East                   & 0.071                          & 0.063                              & 0.721                   \\ \hline
North America                 & 0.254                          & 0.221                              & 0.307                   \\ \hline
Other                         & 0.015                          & 0.013                              & 1.000                   \\ \hline
NA                            & 0.084                          & 0.125                              & 0.070                   \\ \hline
\end{tabular}

\vspace{8mm}

\begin{tabular}{|l|l|l|l|}
\hline
\textbf{Gender} & \textbf{Respondent proportion} & \textbf{Non-respondent proportion} & \textbf{p-test \textit{p}-value} \\ \hline
female          & 0.054                          & 0.100                              & 0.020                   \\ \hline
male            & 0.919                          & 0.842                              & 0.001                   \\ \hline
NA              & 0.027                          & 0.058                              & 0.048                   \\ \hline
\end{tabular}

\vspace{8mm}

\begin{tabular}{|l|l|l|l|l|}
\hline
\textbf{Variable} & \textbf{Respondent estimate} & \textbf{Non-respondent estimate} & \textbf{statistic} & \textbf{\textit{p}-value} \\ \hline
Citations         & 2740.5                       & 4528.0                           & 2.55               & 0.010856         \\ \hline
log(Citations)    & 5.9                          & 6.4                              & 3.19               & 0.001490         \\ \hline
Years in field    & 8.6                          & 11.1                             & 4.04               & 0.000060         \\ \hline
\end{tabular}
\end{table}

\hlinetable

\newpage
\subsection*{S4: Survey responses on AI progress, intelligence explosions, and AI Safety}
Three of the questions below concern Stuart Russell's argument about highly advanced AI. An excerpt of the argument was included in the survey. The full argument can be found here: \\ \url{www.edge.org/conversation/the-myth-of-ai#26015}.

\begin{table}[H]
\centering

\includegraphics[width=\textwidth]{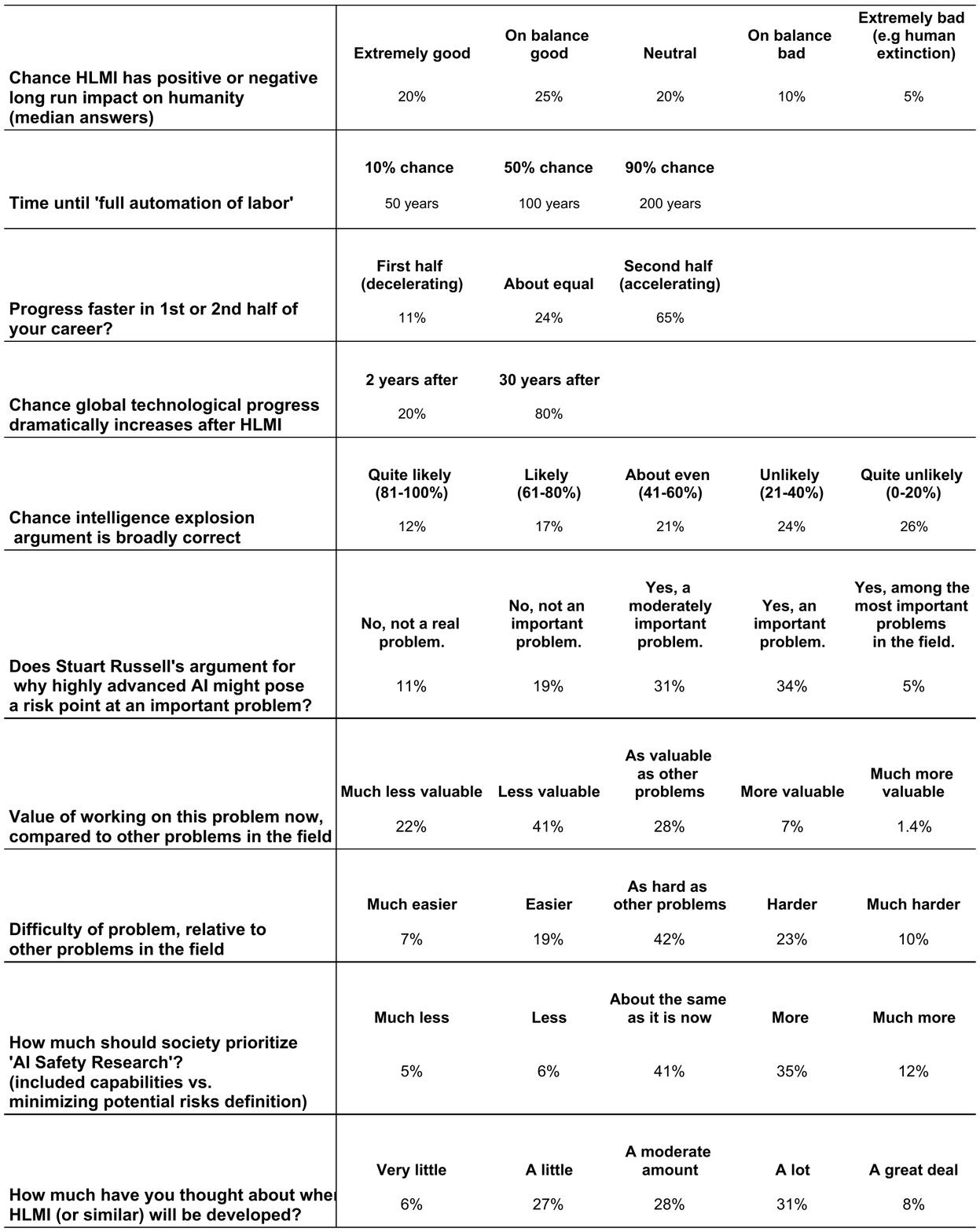}
\caption{Median survey responses for AI progress and safety questions}\label{tab:s4}
\end{table}

\newpage
\subsection*{S5: Description of AI Milestones}
The timelines in Figure \ref{fig:timeline} are based on respondents’ predictions about the achievement of various milestones in AI. Beliefs were elicited in the same way as for HLMI predictions (see ``Elicitation of Beliefs'' above). We chose a subset of all milestones to display in Figure \ref{fig:timeline} based on which milestones could be accurately described with a short label.

\begin{longtable}{| p{4.5cm} | p{4.5cm}  | l | l | p{1.5cm} |}
\caption{\textbf{Descriptions of AI Milestones} }\\
\hline
\textbf{Milestone Name}                                        & \textbf{Description} & \textbf{n} & \textbf{In Fig. \ref{fig:timeline}} & \textbf{median (years)} \\ \hline
Translate New Language with 'Rosetta Stone'                    & Translate a text written in a newly discovered language into English as well as a team of human experts, using a single other document in both languages (like a Rosetta stone). Suppose all of the words in the text can be found in the translated document, and that the language is a difficult one.            & 35         &                   & 16.6                    \\ \hline
Translate Speech Based on Subtitles                            & Translate speech in a new language given only unlimited films with subtitles in the new language. Suppose the system has access to training data for \textit{other} languages, of the kind used now (e.g., same text in two languages for many languages and films with subtitles in many languages).               & 38         &                   & 10                      \\ \hline
Translate (vs. amateur human)                                  & Perform translation about as good as a human who is fluent in both languages but unskilled at translation, for most types of text, and for most popular languages (including languages that are known to be difficult, like Czech, Chinese and Arabic).          & 42         & X                 & 8                       \\ \hline
Telephone Banking Operator                                     & Provide phone banking services as well as human operators can, without annoying customers more than humans. This includes many one-off tasks, such as helping to order a replacement bank card or clarifying how to use part of the bank website to a customer.                & 31         & X                 & 8.2                     \\ \hline
Make Novel Categories                                          & Correctly group images of previously unseen objects into classes, after training on a similar labeled dataset containing completely different classes. The classes should be similar to the ImageNet classes.            & 29         &                   & 7.4                     \\ \hline
One-Shot Learning                                              & One-shot learning: see only one labeled image of a new object, and then be able to recognize the object in real world scenes, to the extent that a typical human can (i.e. including in a wide variety of settings). For example, see only one image of a platypus, and then be able to recognize platypuses in nature photos. The system may train on labeled images of other objects.
%
%
 
Currently, deep networks often need hundreds of examples in classification tasks[1], but there has been work on one-shot learning for both classification[2] and generative tasks[3].

[1] Lake et al. (2015). Building Machines That Learn and Think Like People

[2] Koch (2015) Siamese Neural Networks for One-Shot Image Recognition

[3] Rezende et al. (2016). One-Shot Generalization in Deep Generative Models
                  & 32         &                   & 9.4                     \\ \hline
Generate Video from New Direction                              & See a short video of a scene, and then be able to construct a 3D model of the scene good enough to create a realistic video of the same scene from a substantially different angle. 

For example, constructing a short video of walking through a house from a video taking a very different path through the house.
                  & 42         &                   & 11.6                    \\ \hline
Transcribe Speech                                              & Transcribe human speech with a variety of accents in a noisy environment as well as a typical human can.           & 33         & X                 & 7.8                     \\ \hline
Read Text Aloud (text-to-spech)                                & Take a written passage and output a recording that can’t be distinguished from a voice actor, by an expert listener.                 & 43         & X                 & 9                       \\ \hline
Math Research                                                  & Routinely and autonomously prove mathematical theorems that are publishable in top mathematics journals today, including generating the theorems to prove.            & 31         & X                 & 43.4                    \\ \hline
Putnam Math Competition                                        & Perform as well as the best human entrants in the Putnam competition—a math contest whose questions have known solutions, but which are difficult for the best young mathematicians.              & 45         & X                 & 33.8                    \\ \hline
Go (same training as human)                                    & Defeat the best Go players, training only on as many games as the best Go players have played. 

For reference, DeepMind’s AlphaGo has probably played a hundred million games of self-play, while Lee Sedol has probably played 50,000 games in his life[1].

[1] Lake et al. (2015). Building Machines That Learn and Think Like People
               & 42         & X                 & 17.6                    \\ \hline
Starcraft                                                      &      Beat the best human Starcraft 2 players at least 50

Starcraft 2 is a real time strategy game characterized by:
\begin{itemize}
\item Continuous time play
\item Huge action space
\item Partial observability of enemies
\item Long term strategic play, e.g. preparing for and then hiding surprise attacks.
\end{itemize}

                & 24         & X                 & 6                       \\ \hline
Quick Novice Play at Random Game                               & Play a randomly selected computer game, including difficult ones, about as well as a human novice, after playing the game less than 10 minutes of game time. The system may train on other games.                 & 44         &                   & 12.4                    \\ \hline
Angry Birds                                                    & Play new levels of Angry Birds better than the best human players. Angry Birds is a game where players try to efficiently destroy 2D block towers with a catapult. For context, this is the goal of the IJCAI Angry Birds AI competition.             & 39         & X                 & 3                       \\ \hline
All Atari Games                                                & Outperform professional game testers on all Atari games using no game-specific knowledge. This includes games like Frostbite, which require planning to achieve sub-goals and have posed problems for deep Q-networks[1][2].

[1] Mnih et al. (2015). Human-level control through deep reinforcement learning.

[2] Lake et al. (2015). Building Machines That Learn and Think Like People
           & 38         & X                 & 8.8                     \\ \hline
Novice Play at half of Atari Games in 20 Minutes               &         Outperform human novices on 50\% of Atari games after only 20 minutes of training play time and no game specific knowledge. 

For context, the original Atari playing deep Q-network outperforms professional game testers on 47\% of games[1], but used hundreds of hours of play to train[2].

[1] Mnih et al. (2015). Human-level control through deep reinforcement learning.

[2] Lake et al. (2015). Building Machines That Learn and Think Like People
             & 33         &                   & 6.6                     \\ \hline
Fold Laundry                                                   & Fold laundry as well and as fast as the median human clothing store employee.                 & 30         & X                 & 5.6                     \\ \hline
5km Race in City (bipedal robot vs. human)                     & Beat the fastest human runners in a 5 kilometer race through city streets using a bipedal robot body.                 & 28         & X                 & 11.8                    \\ \hline
Assemble any LEGO                                              & Physically assemble any LEGO set given the pieces and instructions, using non- specialized robotics hardware. 

For context, Fu 2016[1] successfully joins single large LEGO pieces using model based reinforcement learning and online adaptation.

[1] Fu et al. (2016). One-Shot Learning of Manipulation Skills with Online Dynamics Adaptation and Neural Network Priors
           & 35         & X                 & 8.4                     \\ \hline
Learn to Sort Big Numbers Without Solution Form                & Learn to efficiently sort lists of numbers much larger than in any training set used, the way Neural GPUs can do for addition[1], but without being given the form of the solution. 

For context, Neural Turing Machines have not been able to do this[2], but Neural Programmer-Interpreters[3] have been able to do this by training on stack traces (which contain a lot of information about the form of the solution).

[1] Kaiser \& Sutskever (2015). Neural GPUs Learn Algorithms 

[2] Zaremba \& Sutskever (2015). Reinforcement Learning Neural Turing Machines 

[3] Reed \& de Freitas (2015). Neural Programmer-Interpreters
                & 44         &                   & 6.2                     \\ \hline
Python Code for Simple Algorithms                              & Write concise, efficient, human-readable Python code to implement simple algorithms like quicksort. That is, the system should write code that sorts a list, rather than just being able to sort lists. 

Suppose the system is given only:

\begin{itemize}
\item A specification of what counts as a sorted list
\item Several examples of lists undergoing sorting by quicksort
\end{itemize}
                & 36         &                   & 8.2                     \\ \hline
Answer Factoid Questions via Internet                          & Answer any ``easily Googleable'' factoid questions posed in natural language better than an expert on the relevant topic (with internet access), having found the answers on the internet. 

Examples of factoid questions:

\begin{itemize}
\item ``What is the poisonous substance in Oleander plants?'' 
\item ``How many species of lizard can be found in Great Britain?''
\end{itemize}
               & 46         &                   & 7.2                     \\ \hline
Answer Open-Ended Factual Questions via Internet               & Answer any ``easily Googleable'' factual but open ended question posed in natural language better than an expert on the relevant topic (with internet access), having found the answers on the internet. 

Examples of open ended questions: 
\begin{itemize}
\item ``What does it mean if my lights dim when I turn on the microwave?'' 
\item ``When does home insurance cover roof replacement?"
\end{itemize}

               & 38         &                   & 9.8                     \\ \hline
Answer Questions Without Definite Answers                      & Give good answers in natural language to factual questions posed in natural language for which there are no definite correct answers. 

For example: ``What causes the demographic transition?'', ``Is the thylacine extinct?'', ``How safe is seeing a chiropractor?''
                 & 47         &                   & 10                      \\ \hline
High School Essay                                              & Write an essay for a high-school history class that would receive high grades and pass plagiarism detectors. 

For example answer a question like ``How did the whaling industry affect the industrial revolution?''
                & 42         & X                 & 9.6                     \\ \hline
Generate Top 40 Pop Song                                       & Compose a song that is good enough to reach the US Top 40. The system should output the complete song as an audio file.              & 38         & X                 & 11.4                    \\ \hline
Produce a Song Indistinguishable from One by a Specific Artist & Produce a song that is indistinguishable from a new song by a particular artist, e.g., a song that experienced listeners can't distinguish from a new song by Taylor Swift.              & 41         &                   & 10.8                    \\ \hline
Write New York Times Best-Seller                               & Write a novel or short story good enough to make it to the New York Times best-seller list.                & 27         & X                 & 33                      \\ \hline
Explain Own Actions in Games                                   & For any computer game that can be played well by a machine, explain the machine’s choice of moves in a way that feels concise and complete to a layman.                  & 38         & X                 & 10.2                    \\ \hline
World Series of Poker                                          & Play poker well enough to win the World Series of Poker.                 & 37         & X                 & 3.6                     \\ \hline
Output Physical Laws of Virtual World                          & After spending time in a virtual world, output the differential equations governing that world in symbolic form. 

For example, the agent is placed in a game engine where Newtonian mechanics holds exactly and the agent is then able to conduct experiments with a ball and output Newton's laws of motion.
                & 52         &                   & 14.8                    \\ \hline
\label{tab:s5}
\end{longtable}

\newpage
\subsection*{Acknowledgments}
We thank Connor Flexman for collecting demographic information. We also thank Nick Bostrom for inspiring this work, and Michael Webb and Andreas Stuhlm\"uller for helpful comments. We thank the Future of Humanity Institute (Oxford), the Future of Life Institute, and the Open Philanthropy Project for supporting this work.

\end{document}